\documentclass[letterpaper, 10 pt, conference]{ieeeconf}  

\IEEEoverridecommandlockouts                              

\title{\LARGE \bf
Revisiting Deformable Convolution for Depth Completion
}

\author{Xinglong Sun$^{1}$, Jean Ponce$^{2}$, and Yu-Xiong Wang$^{3}$
\thanks{$^{1}$Xinglong Sun is a graduate student in the Department of Computer Science at Stanford University,
        {\tt xs15@stanford.edu}}%
\thanks{$^{2}$Jean Ponce is a Professor in the Computer Science Department of Ecole normale supérieure  (ENS-PSL, CNRS, Inria) and a Global Distinguished Professor in the Courant Institute and the Center for Data Science at New York University,
        {\tt jean.ponce@ens.fr}}%
\thanks{$^{3}$Yu-Xiong Wang is an assistant professor in the Department of Computer Science at University of Illinois Urbana-Champaign,
        {\tt yxw@illinois.edu}}%
}

\usepackage{graphicx}
\usepackage{amsmath}
\usepackage{amssymb}
\usepackage{booktabs, makecell, tabularx}
\usepackage[table, dvipsnames]{xcolor}

\usepackage[pagebackref,breaklinks,colorlinks]{hyperref}
\usepackage{algorithm}
\usepackage{algorithmicx}
\usepackage{algpseudocode}
\usepackage{cite}
\let\labelindent\relax
\usepackage{enumitem}
\usepackage{adjustbox}
\usepackage{newfloat}
\usepackage[utf8]{inputenc} 
\usepackage[T1]{fontenc}
\usepackage{listings}
\usepackage{bibentry}
\usepackage{arydshln}
\usepackage{pifont}

\usepackage[capitalize]{cleveref}
\usepackage{bbold}
\usepackage{multirow}

\usepackage{etoolbox}
\makeatletter
\patchcmd{\@makecaption}
  {\scshape}
  {}
  {}
  {}
\makeatother

\usepackage{footnote}
\usepackage{url}
\makesavenoteenv{tabular}
\makesavenoteenv{table}

\begin{document}

\newcommand\method{Ours}
\maketitle
\thispagestyle{empty}
\pagestyle{empty}
\begin{abstract}
Depth completion, which aims to generate high-quality dense depth maps from sparse depth maps, has attracted increasing attention in recent years. Previous work usually employs RGB images as guidance, and introduces iterative spatial propagation to refine estimated coarse depth maps. However, most of the propagation refinement methods require several iterations and suffer from a fixed receptive field, which may contain irrelevant and useless information with very sparse input. In this paper, we address these two challenges simultaneously by revisiting the idea of deformable convolution. We propose an effective architecture that leverages deformable kernel convolution as a single-pass refinement module, and empirically demonstrate its superiority. To better understand the function of deformable convolution and exploit it for depth completion, we further systematically investigate a variety of representative strategies. Our study reveals that, different from prior work, deformable convolution needs to be applied on an estimated depth map with a relatively high density for better performance. We evaluate our model on the large-scale KITTI dataset and achieve state-of-the-art level performance in both accuracy and inference speed. Our code is available at \href{https://github.com/AlexSunNik/Revisiting-Deformable-Convolution-for-Depth-Completion/tree/main}{\textcolor{blue}{https://github.com/AlexSunNik/ReDC}}.

\end{abstract}

\section{INTRODUCTION}
Reliable depth information is fundamental in many real-world applications, such as robotics~\cite{shankar2022learned}, autonomous driving~\cite{natan2022end}, and 3D mapping~\cite{hervier2012accurate}. It is critical for agents to better sense the surrounding environment for more accurate and effective actions. However, existing depth sensors usually produce depth maps with incomplete data. For example, LiDAR sensors have a limited number of scanlines and operating frequencies, thus producing very sparse depth measurements. Therefore, depth completion, which aims to estimate dense depth maps from the sparse ones, has received increasing attention recently.\par
A wide variety of deep convolutional neural network~\cite{lecun1998gradient} based approaches have been developed for effective depth completion. They often formulate the task as a learned interpolation, using high-quality RGB images from camera sensors as guidance to up-sample sparse points into dense depth images. Inspired by monocular depth estimation~\cite{li2018megadepth, fu2018deep, pan2019method, kumari2019autodepth}, these approaches~\cite{ma2018sparse, shivakumar2019dfusenet, uberatg, zhao2021adaptive, jaritz2018sparse, uhrig2017sparsity, eldesokey2019confidence, qiu2019deeplidar, xu2019depth, liu2023mff, liang2022selective} typically use an encoder-decoder network. In addition, to better leverage the two different modalities of color and depth data, two-branch network architectures are adopted. \par

\begin{figure*}
\begin{center}
\includegraphics[width=.8\linewidth]{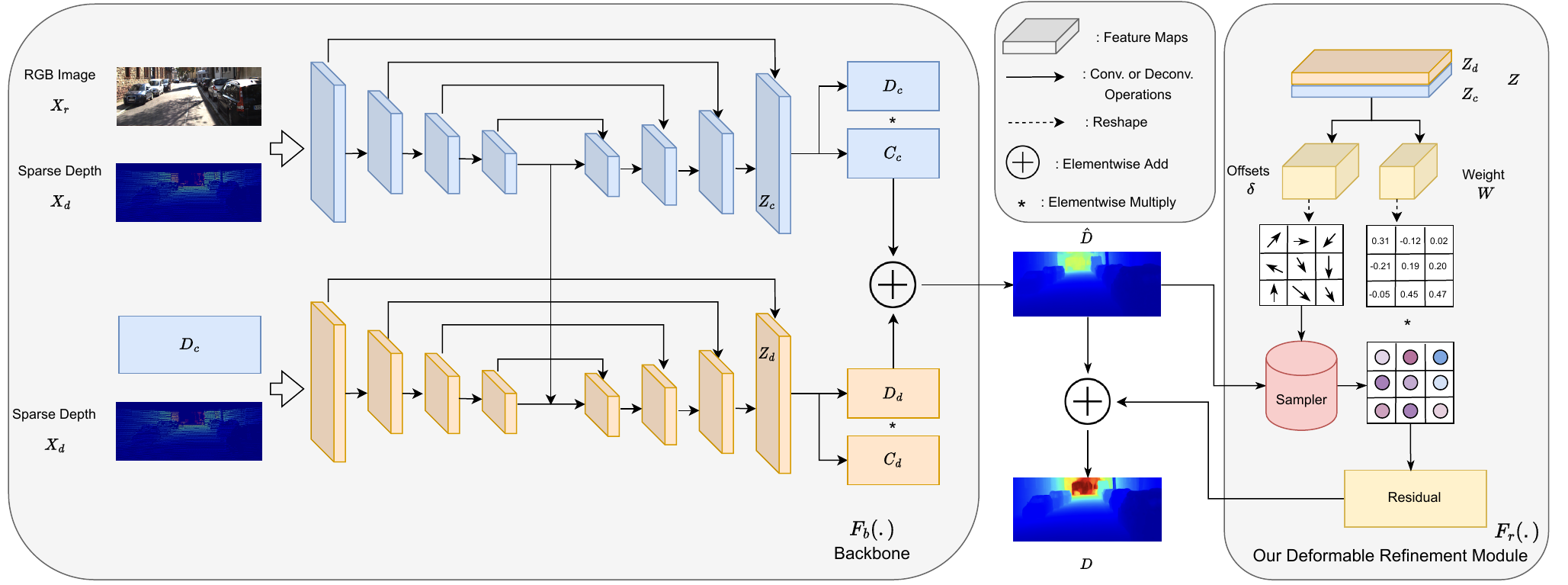}
\end{center}
\vspace{-4mm}
  \caption{Overview of our method ReDC. We first generate a coarse depth map $\hat{D}$ from the backbone. Then, we pass it through our deformable refinement module. We feed the feature maps coming from the final layer of the backbone decoder into a $1\times 1$ convolution layer to regress a set of weights and offsets for performing the interpolation on $\hat{D}$ in a deformable manner.}
  \vspace{-5mm}
\label{fig:flowchart}
\end{figure*}

Nonetheless, the depth maps directly output from the decoder are often still blurry or low in quality. Concretely, as observed in~\cite{cheng2018depth}, depth maps directly produced by deep decoders may not preserve the input depth values at valid pixels. Many approaches~\cite{cheng2018depth, cheng2019learning, cheng2020cspn++, park2020non, xu2020deformable} thus introduce efficient spatial propagation to refine coarse depth maps. However, most of the propagation refinement methods suffer from two major drawbacks. First, propagation at each pixel is performed with a uniform-size receptive field and a regular convolution. This limitation originates from the fixed geometric structure of the convolution module. A convolution unit can only sample the input feature map at fixed locations, which may contain {\em irrelevant and useless information with very sparse input}. Second, the propagation refinement module usually needs to refine the coarse depth map for {\em several iterations} to obtain satisfactory performance, which is not ideal for good hardware latency and memory footprint.

In this paper, we address these two challenges simultaneously by \textbf{Re}visiting the idea of {\em \textbf{D}eformable \textbf{C}onvolution}~\cite{dai2017deformable} (\textbf{ReDC}), considering its capability of adaptively generating different receptive fields by learning offsets on regular grid positions. {\em Our key insight} is that such a capability (i) is desirable for depth completion so as to adaptively attend to the most informative regions, and (ii) should be exploited to refine coarse depth maps. To this end, we propose an effective and efficient architecture that leverages deformable kernel convolution as a refinement module and empirically demonstrate its superiority. We build upon the PENet\cite{hu2021penet} backbone, one of the publicly released state-of-the-art methods. We observe a significant improvement over PENet with our architecture. Notably, we found that with our deformable refinement module, the coarse depth {\em only needs a single pass of refinement}, distinguishing our model from those iterative ones~\cite{cheng2020cspn++, xu2020deformable, park2020non, hu2021penet} and enjoying noticeable inference speedup. 

Our approach is also different from and substantially outperforms prior work that uses deformable convolution for depth completion tasks~\cite{xu2020deformable, park2020non, kim2019deformable}, whose performance is still inferior to the state of the art. For example, Deformable Kernel Network (DKN)~\cite{kim2019deformable} obtains strong performance on the NYU-v2 benchmark~\cite{Silberman:ECCV12}, but it performs poorly on the more challenging KITTI benchmark~\cite{Geiger2013IJRR, uhrig2017sparsity}, where the input depth maps exhibit much higher sparsity. Similarly, DSP~\cite{xu2020deformable} and NLSPN~\cite{park2020non} fail to achieve competitive performance on KITTI, despite several iterations of refinement.

To better understand the function of deformable convolution and leverage it for depth completion, we systematically investigate a variety of representative strategies other than our proposed approach. Our study reveals that different ways of employing deformable convolution lead to significantly different performance. {\em Importantly, it needs to be applied on a depth map with a relatively high density for better performance}. In other words, a direct application of deformable convolution as in DKN\cite{kim2019deformable} does not work with high input depth sparsity like KITTI. This further supports our proposed method of using deformable refinement over coarse depth.

We summarize our contributions as follows:
\begin{itemize}
    \item We conduct a thorough study on the most effective way of leveraging deformable convolution, promoting a deeper understanding of deformable convolution within the depth completion research community.
    \item We propose a novel and effective depth completion architecture that requires only a single pass through the depth refinement module, achieving more hardware and memory-friendly performance.
    \item We achieve competitive state-of-the-art level depth completion performance in accuracy and inference speed on the challenging KITTI dataset and surpass prior work by a clear margin.
\end{itemize}
\section{RELATED WORK}
\label{sec:related}
\subsection{Depth Completion}
Depth completion aims to generate dense depth maps by completing sparse depth maps. Some work~\cite{uhrig2017sparsity, eldesokey2020uncertainty} produces dense depth maps from monocular sparse depth maps only, while the majority of work~\cite{hua2018normalized, qiu2019deeplidar, liu2013guided, cheng2018depth} employs guidance or reference images like high-quality RGB images to assist the dense depth map generation. These approaches try to solve the challenge of irregular and extremely sparse input depth and the two completely different modalities of color and depth with sparse and spatially-invariant convolution~\cite{hua2018normalized, uhrig2017sparsity, eldesokey2018propagating,  huang2019hms}, uncertainty exploration~\cite{van2019sparse, eldesokey2019confidence}, and model fusion strategies~\cite{tang2020learning}. Additionally, some representative work tries to draw insights from other tasks and exploits multi-scale features~\cite{huang2019hms, eigen2014depth, wang2018multi, li2020multi}, surface normals~\cite{qiu2019deeplidar, xu2019depth}, semantic information~\cite{jaritz2018sparse, schneider2016semantically}, and context affinity~\cite{cheng2020cspn++, cheng2018depth, park2020non} to improve depth completion performance. Recently, PENet is proposed and incorporates several effective features useful to boost depth completion performance. Moreover, it proposes a simple convolutional layer to encode 3D geometric cues which are shown to further improve the performance. PENet is one of the best publicly released models on the KITTI depth completion benchmark, and we study our deformable refinement module upon it and observe a clear improvement. 

\subsection{Deformable Convolution}
The idea of deformable convolution is first introduced and explored in DCN~\cite{dai2017deformable} to enhance the transformation modeling capability of convolutional neural networks (CNNs). Due to the intrinsic nature of the depth completion task, deformable convolution is quickly exploited to extract more useful information from the very sparse depth maps. DKN~\cite{kim2019deformable} explicitly learns sparse and spatially-invariant kernels for depth completion by regressing offsets to sample pixels at adaptive locations. It obtains superior performance on the NYU-v2 depth~\cite{Silberman:ECCV12} completion benchmark, indicating the potential effectiveness of the DKN module on the depth completion task. However, our investigation shows that its performance is degraded on the more challenging KITTI benchmark with input depth of much higher sparsity. DSP~\cite{xu2020deformable} adaptively learns a different receptive field and affinity matrix between non-local pixels by generating memory-costly embedding for each non-local pixel, which is impractical on edge devices. Still, it fails to reach state-of-the-art level performance on KITTI and requires a minimum of $3$ iterations of refinement for decent performance, which is also the case in another deformable model NLSPN~\cite{park2020non}. We conduct a thorough study on deformable convolution and explore its more effective usage with very sparse input depth maps like KITTI. We propose a new efficient architecture which achieves state-of-the-art level performance while using much less inference time, and we comprehensively analyze its effectiveness through experiments.
\section{METHODOLOGY}
\label{sec:method}

We design an end-to-end trainable model ReDC with our proposed deformable refinement module, as illustrated in Fig.~\ref{fig:flowchart}. We first feed a sparse depth map and a guidance RGB image into the backbone model, which produces a coarse depth map. We then use our refinement module based on deformable kernel convolution to improve its quality. As mentioned previously, we only pass it through the refinement module \emph{once}.
\subsection{Overall Procedure}
We represent the input sparse depth map as $X_d$, the guidance RGB image as $X_r$, and the provided groundtruth dense map in the training set as $Y$. We denote our completion model as $F(.)$ which produces a dense depth map $D$ from $X_d$ and $X_r$. Concretely, this is expressed as $D = F(X_d, X_r)$. The completion model contains two parts, namely, the backbone $F_b(.)$ which produces a coarse depth map $\hat{D}$ and the refinement module $F_r(.)$ which improves the quality of the coarse depth. The overall model can be expressed as:
\begin{align}
    D = F_r(F_b(X_d, X_r)).
\end{align}
We next discuss the backbone choice and our deformable refinement module.
\subsection{Backbone}
We follow PENet\cite{hu2021penet} for the backbone choice. The backbone network has a two-branch architecture, with one branch extracting the color-dominant information and the other extracting the depth-dominant information. Each branch is of an encoder-decoder structure, and the results coming from both branches are fused to generate the coarse depth $\hat{D}$. Depth generated by the color-dominant branch is denoted as $D_c$, and depth from the depth-dominant branch is denoted as $D_d$. For fusing the two depth maps, confidence masks are also generated, which are $C_c$ for $D_c$ and $C_d$ for $D_d$. We perform the fusion following PENet and FusionNet\cite{van2019sparse}. Concretely, the final fused depth $\hat{D}$ is computed as:
\begin{align}
    \hat{D}(u,v) = \frac{D_d(u,v)\cdot e^{C_d(u,v)}+D_c(u,v)\cdot e^{C_c(u,v)}}{e^{C_d(u,v)}+e^{C_c(u,v)}},
\end{align}
where $(u,v)$ denotes the location of each pixel. In practice, a softmax is applied on $C_c$ and $C_d$ for normalization.
\subsection{Refinement Module}
\label{subsec:refinement}
We now present our deformable refinement module in detail and how we utilize it to improve the quality of the generated depth. Following previous work~\cite{he2012guided, kopf2007joint, tomasi1998bilateral, wu2018fast, kim2019deformable}, we represent each point of our refined depth $D$ as a weighted average of points sampled from the coarse depth $\hat{D}$. Specifically, we represent it as:
\begin{align}
D_p = \sum_{q\in \mathcal{N}(p)}W_{pq}(Z)\hat{D}_q,    
\end{align}
where $p=(u,v)$ represents the point coordinate on the depth map, $\mathcal{N}(p)$ defines a set of neighbors near the point $p$, and $Z$ is some guidance like extracted feature maps. Most of the approaches~\cite{hui2016depth, li2016deep, li2019joint, cheng2020cspn++} mainly focus on learning the weights $W$ and employ a fixed square grid as predefined neighbors to sample points. Deformable approaches~\cite{dai2017deformable, xu2020deformable, kim2019deformable} try to learn a set of offsets added to each sampled location in the regular square grid, producing a deformable neighborhood region. 

Here, we show how we design the architecture to learn the kernel weights $W$ and neighborhood $\mathcal{N}$ offsets.
We denote the feature map coming from the last layer of the decoder of the color-dominant branch as $Z_c$ and that from the depth-dominant branch as $Z_d$, as illustrated in Fig. \ref{fig:flowchart}. We concatenate the two feature maps channel-wise and represent the combined feature as $Z=\{Z_d, Z_c\}$. Later, we add a $1 \times 1$ convolutional layer taking the combined feature $Z$ as input. It outputs a feature map of size $k^2 \times 1 \times 1$, where $k$ represents the size of the interpolation kernel $W$. We then apply $sigmoid(.)$ on it to estimate the interpolation kernel weights. Similar to ~\cite{kim2019deformable}, we found that a softmax instead of sigmoid layer as used in ~\cite{bako2017kernel, niklaus2017video, vogels2018denoising} yields no performance improvement. Same as the weights, we add another $1\times1$ convolutional layer on feature $Z$ to regress the offsets as well. It produces an output feature map of size $2k^2\times1\times1$ containing the relative offsets (both in $x$ and $y$ directions) we use to apply to sampled locations on a regular grid. For example, with $k=3$, we would learn $9$ offsets used to sample $9$ points sparsely chosen from a large neighborhood area. Similar to \cite{kim2019deformable}, we choose to compute the residual instead of directly generating the final output. The refinement of the coarse depth $\hat{D}$ can thus be expressed as follows:
\begin{align}
    D_p = \hat{D}_p + \sum_{q\in\mathcal{N}(p)}W_{ps}(Z)\hat{D}_{s(q)},
\end{align}
where $\mathcal{N}(p)$, as previously described, is a local $k\times k$ regular square window centered at location $p$. We express $s(q)$ the sampled position as $s(q) = q + \delta q$, where $\delta q$ comes from the $2k^2\times1\times1$ shaped offset map $\delta$ we regress. Since $s(q)$ could be fractional, in order to compute $\hat{D}_{s(q)}$, we utilize a two-dimensional bilinear kernel as in~\cite{dai2017deformable, jaderberg2015spatial} to compute its value based on its $4$ neighbors, detailed as follows:
\begin{align}
    \hat{D}_{s(q)} = \sum_{t\in\mathcal{R}(s(q))}G(s(q),t)\hat{D}_{t},
\end{align}
where $\mathcal{R}(s(q))$ lists all of the four integer neighbors of $s(q)$ and $G$ is a sampling kernel defined as:
\begin{align}
    G(s,t) = \mathrm{max}(0, 1-|s_x-t_x|)\cdot \mathrm{max}(0, 1-|s_y-t_y|).
\end{align}

\begin{figure*}[t!]
\begin{center}
\includegraphics[width=\linewidth]{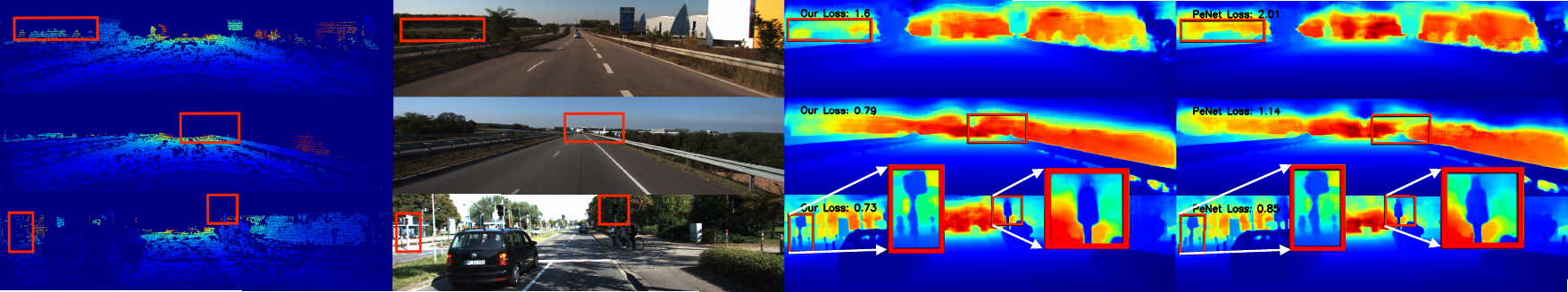}
\end{center}
\vspace{-4mm}
  \caption{Visualization of inference results obtained by our model ReDC and PENet\cite{hu2021penet} on three challenging samples from the KITTI {\em training} set. From \textbf{left} to \textbf{right}: groundtruth dense depth map, guidance RGB image, inference from our model, and inference from PENet. $\ell_2$ loss computed between the groundtruth and inference result is presented at the top left corner. Regions to focus on are highlighted in red boxes, showing that our ReDC achieves better performance with more accurate depth and finer detail.}
  \vspace{-3mm}
\label{fig:train_worst}
\end{figure*}

\begin{figure*}[t!]
\begin{center}
\includegraphics[width=\linewidth]{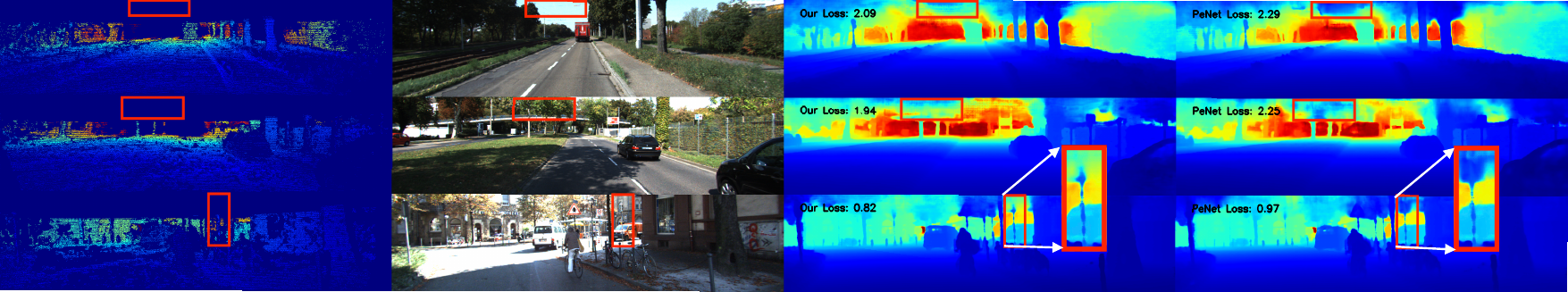}
\end{center}
\vspace{-4mm}
  \caption{Visualization of inference results obtained by our model ReDC and PENet\cite{hu2021penet} on three challenging samples from the KITTI {\em validation} set. Notation follows Figure~\ref{fig:train_worst}. Consistent with Figure~\ref{fig:train_worst}, our ReDC achieves better performance with more accurate depth and finer detail.}
  \vspace{-3mm}
\label{fig:val_worst}
\end{figure*}

\subsection{Training Loss}
We use a combined $\ell_2$ and $\ell_1$ loss for training our model, which is defined as:
\begin{align}
\begin{split}
    L(D, Y) = \alpha \|(D - Y) \odot\mathbb{1}(Y>0) \|^2 + \\ (1-\alpha) \|(D - Y) \odot\mathbb{1}(Y>0) \|.
\end{split}
\end{align}
Here, $\odot$ is the elementwise multiplication. For datasets like KITTI, even the groundtruth depth map is of relatively high sparsity, containing many invalid pixels. We follow the common practice and only consider those having valid depth values with the mask $\mathbb{1}(Y>0)$.

\section{EXPERIMENTS}
\label{sec:exp}

\subsection{Experiment Settings}
\label{subsec:setting}

\textbf{Dataset:}
\label{subsec:dataset}
We evaluate our approach on the KITTI~\cite{Geiger2013IJRR} Depth Completion benchmark~\cite{uhrig2017sparsity}, with sparse depth maps obtained by projecting 3D LiDAR points to corresponding image frames. The density of the valid depth points is around $5\%$ for the sparse depth maps and $16\%$ for the groundtruth dense depth maps. The dataset contains $93K$ data samples, with $86K$ for training, $7K$ for validation, and $1K$ for testing.

\textbf{Evaluation Metrics:}
\label{subsubsec:metrics}
We follow the common practice and evaluate our approach under four standard metrics~\cite{uhrig2017sparsity}, including root mean square error (RMSE) in $mm$, mean absolute error (MAE) in $mm$, root mean squared error of the inverse depth (iRMSE) in $1/km$, and mean absolute error of the inverse depth (iMAE) in $1/km$. We also report the inference time for a single depth-image completion in seconds on RTX 2080Ti GPU with a 2.5GHz processor.

\textbf{Baselines:}
\label{subsubsec:baselines}
We mainly compare with competitive {\em publicly} released and \emph{peer-reviewed} depth completion methods, including PENet~\cite{hu2021penet} and DSP~\cite{xu2020deformable}. In addition to the discussed methods in Sec. \ref{sec:related}, we include very recent baselines like MFF-Net~\cite{liu2023mff}, RigNet~\cite{yan2022rignet}, and GAENet~\cite{chen2022depth}, as well as well-referenced baselines like PwP~\cite{xu2019depth} and UberATG~\cite{uberatg}. Finally, we evaluate alternative strategies of using deformable convolution for depth completion, such as DKN~\cite{kim2019deformable}.

\textbf{Implementation Detail:}
\label{subsubsec:implement}
We implement our approach and conduct all experiments using PyTorch~\cite{paszke2019pytorch} with four NVIDIA RTX 2080Ti GPUs running in parallel. We train with an Adam optimizer~\cite{kingma2015adam} with weight decay set to $1e-5$, $\beta_1$ set to $0.9$, and $\beta_2$ set to $0.99$. We run the experiment for a total of $30$ epochs with an initial learning rate set to $1e-3$, decaying in a cosine annealing fashion. The batch size is set to $2$ samples per GPU for each step. We choose $k=3$ which is the size of the interpolation kernel $W$ discussed in Section \ref{subsec:refinement}.  The loss function is the combined $\ell_2$ and $\ell_1$ losses with $\alpha$ set to $0.5$. In terms of training data augmentation, we follow the common practice and adopt random cropping, random flipping, and color jittering~\cite{devries2017improved, krizhevsky2017imagenet}.

\subsection{Main Results}
\label{subsec:results}

\textbf{Quantitative Comparisons:}
\label{subsubsec:quan}
\begin{figure}[t!]
\begin{center}
\includegraphics[width=\linewidth]{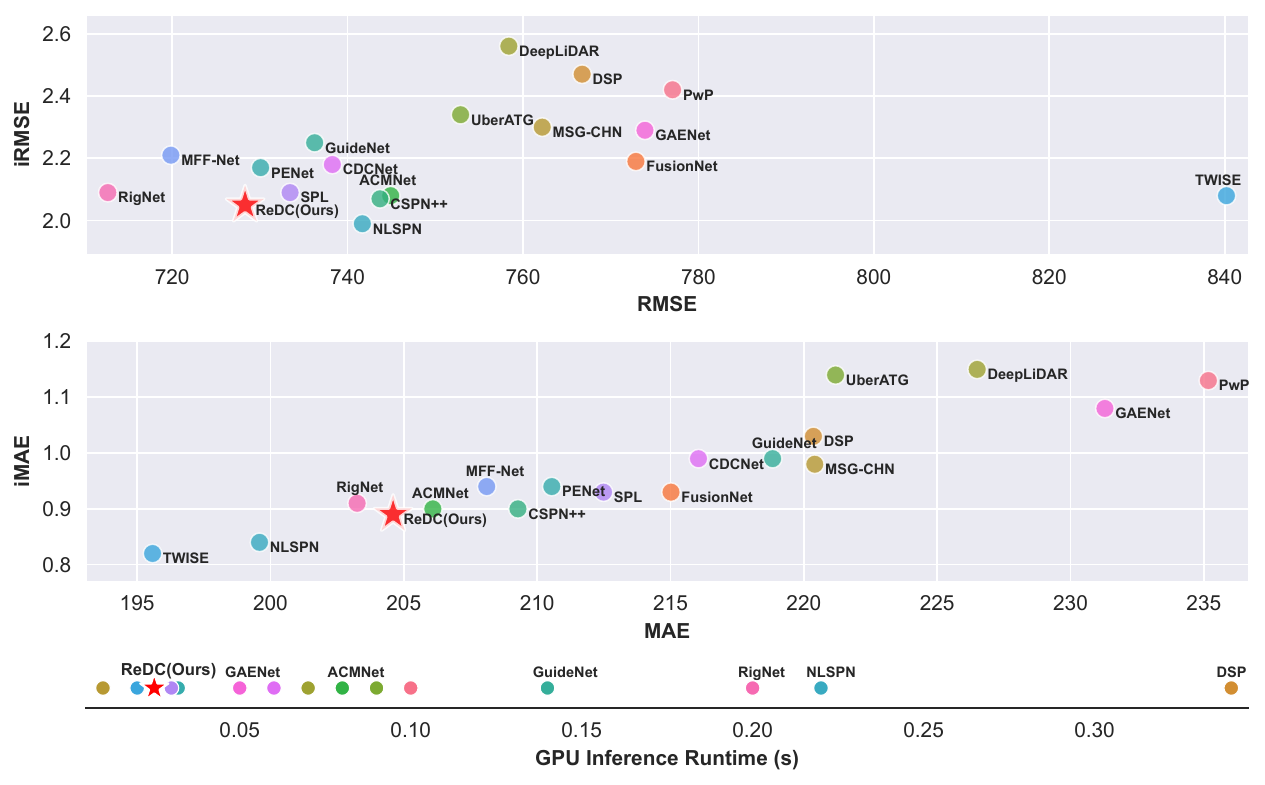}
\end{center}
\vspace{-4mm}
  \caption{Our model ReDC clearly outperforms competitive publicly released methods in terms of accuracy metrics and inference runtime on the KITTI test set. All metrics are the lower the better. Numerically, our ReDC achieves $728.31$ (RMSE), $204.60$ (MAE), $2.05$ (iRMSE), and $0.89$ (iMAE).}
\vspace{-6mm}
\label{fig:results}
\end{figure}

Figure \ref{fig:results} summarizes the comparisons with baselines on the KITTI test set. Notice that all methods are trained and tuned on the \emph{KITTI released training and validation splits} using sparse and dense depth maps of the \emph{provided default density levels}. In terms of the inference runtime, we only include the reported results from baselines \emph{measured on GPU for a fair comparison}. Our method ReDC achieves \emph{state-of-the-art} level performance in both \emph{accuracy} (RMSE, MAE, iRMSE, and iMAE) and \emph{inference runtime}. Notably, our approach clearly outperforms PENet, which our backbone is based on (Sec.~\ref{sec:method}). This result validates the effectiveness of our deformable refinement module. Moreover, we include the comparison with methods like DSP and NLSPN which also use deformable convolution as part of their depth completion models. In addition to the superior accuracy, our model significantly surpasses them in terms of inference speed, demonstrating the strength of our architecture.

\textbf{Visualizations:}
\label{subsubsec:visul}
We now visually compare the generated depth maps from our model and the top-performing baseline PENet in Figures~\ref{fig:train_worst} and~\ref{fig:val_worst}. Note that, while the quantitative metrics (Section~\ref{subsubsec:metrics}) are computed only over the pixel locations with valid groundtruth depth values, these visualizations provide more comprehensive depth completion results over {\em all} pixels. Therefore, they are complementary to demonstrate the superiority of our approach.

In Figure \ref{fig:train_worst}, we show the three most challenging scenes from the training set for both PENet and our method. We visualize the groundtruth depth maps, RGB images, and inference results from PENet and our model. We also provide the $\ell_2$ loss computed between the groundtruth depth maps and inference results at the top left corner as an informative indicator of the quality of the generated depth maps. We highlight some challenging regions of interest in red boxes, where our model performs substantially better than PENet. We discuss the observation in detail as follows.

(I) {\em Ours outperforms PENet on points with groundtruth.} In the highlighted box of the first hard training sample, we see valid depth points in the groundtruth depth map. Our generated depth correctly predicts the depth values in this region, with the same visualization color as the groundtruth. By contrast, PENet produces incorrect depth values.

(II) {\em Ours also outperforms PENet on points without groundtruth.} In the highlighted box of the second hard training sample, we see that there are completely no valid data in the provided groundtruth depth map. For reasonable estimation in this case, the models have to rely on the RGB image or leverage the depth data from informative neighborhood. Our model correctly estimates the depth values in this region and maintains smoothness with the surrounding area. This is attributed to our deformable refinement module, which is able to {\em adaptively attend to the informative regions}. However, PENet fails on such a challenging region, producing a very sharp transition and gap in the middle.

(III) {\em Ours produces depth with finer detail and structure.} In the highlighted box of the third sample, we observe thin traffic light poles in the image. Our model produces their depth with finer detail and structure than that of PENet.

We also provide the same visualization analysis on the validation set in Figure~\ref{fig:val_worst}. We observe the similar phenomenon that our model leads to more accurate depth estimation in challenging regions which contain {\em very few or a limited number of valid depth points} in groundtruth images, and produces object depth with finer detail and structure. For example, in the highlighted box of the third hard validation example, we see that our model generates the depth of a tall traffic pole with very high quality; whereas, PENet produces a bunch of messy point clouds around the top of the pole.

\subsection{Investigation on Variants of Our Model}
\label{subsec:variants}

\begin{figure}[t!]
\begin{center}
\includegraphics[width=.85\linewidth]{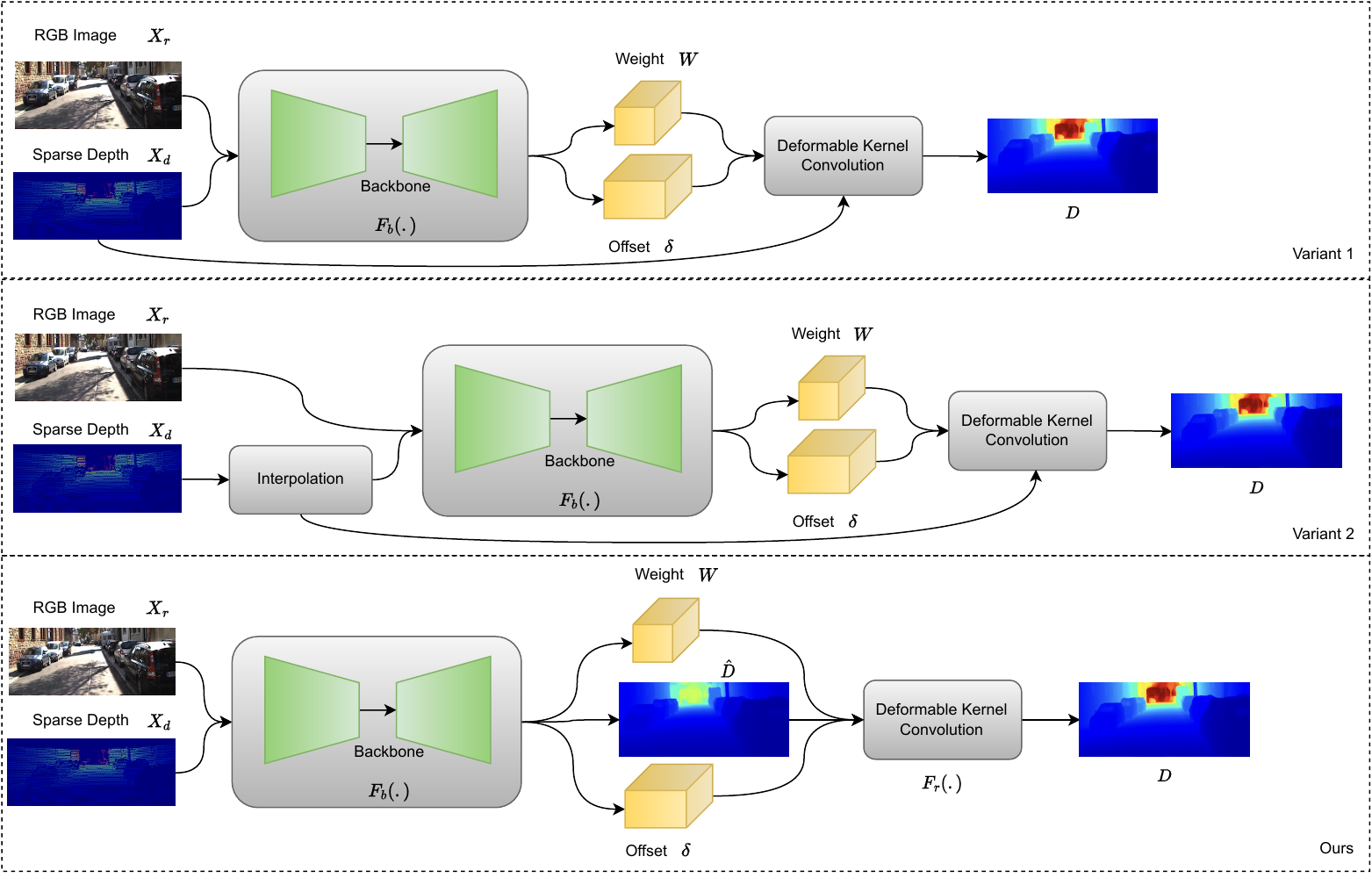}
\end{center}
\vspace{-4mm}
  \caption{Illustration of variants of leveraging deformable convolution for depth completion. Our strategy works the best, which generates an intermediate coarse depth map and then refines it by our deformable refinement module.}

\label{fig:ablation}
\end{figure}

\begin{table}[t!]
\vspace{-2mm}
    \centering
    \caption{Investigation on variants of leveraging deformable convolution for depth completion on the KITTI validation set, demonstrating the superiority of our approach. `NN' denotes nearest-neighbor.}
    \vspace{-3mm}
    \label{table:ablation}
    \resizebox{0.47\textwidth}{!}
    {
        \begin{tabular}{c|cccc}
            \hline
            Model  & RMSE ($\downarrow$) & MAE ($\downarrow$) & iRMSE ($\downarrow$) & iMAE ($\downarrow$)\\\hline
             Variant 1 w/ smaller backbone~\cite{kim2019deformable} & 1130.47 &241.58 &4.00 &0.97\\
             Variant 1 w/ our backbone & 900.54 &251.32 &2.59 &0.99\\
             Variant 2 w/ NN interpolation & 875.75 & 235.58 & 2.40 &0.98\\
            \rowcolor{lightgray} ReDC (Ours) &\textbf{813.05} &\textbf{218.30} &\textbf{2.32} &\textbf{0.89}\\
            \hline
        \end{tabular}
    }
    \vspace{-2mm}
\end{table}

\begin{table}[t!]
    \centering
    \caption{Investigation on different interpolation techniques. Results are reported as directly testing interpolated depth on the KITTI validation set without any model learning. Nearest-neighbor interpolation performs the best overall.}
    \vspace{-3mm}
    \label{table:interpolation}
    \resizebox{0.45\textwidth}{!}
    {
        \begin{tabular}{c|cccc}
            \hline
            Interpolation  & RMSE ($\downarrow$) & MAE ($\downarrow$) & iRMSE ($\downarrow$) & iMAE ($\downarrow$)\\\hline
             RBF & \textbf{2014.52} & 794.89 & inf\footnote{iRMSE and iMAE are inverse metrics. They attain an `inf' (infinite) value, when a model generates $0$ in output depth at valid pixels.} & inf\\
             Linear & 2265.33 & 507.93 & inf & inf\\
             Cubic & 3022.93 & 644.99 & inf & inf\\
             Nearest & 2139.25 & \textbf{454.42} & \textbf{6.45} & \textbf{1.83}\\
            \hline
        \end{tabular}
    }
    \vspace{-6mm}
\end{table}

\begin{figure*}
\begin{center}
\includegraphics[width=.95\linewidth]{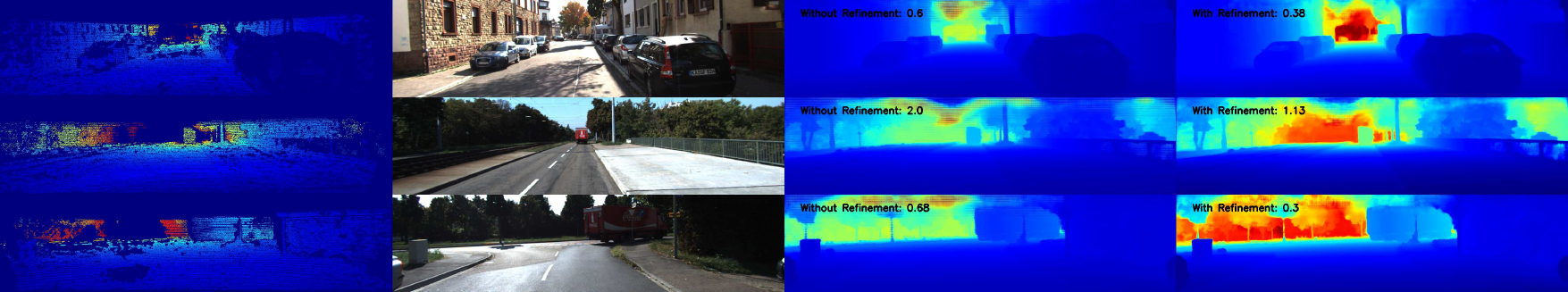}
\vspace{-5.5mm}
\end{center}
  \caption{Visualization of inference results obtained by our model ReDC before and after the deformable refinement module. From \textbf{left} to \textbf{right}: groundtruth dense depth, guidance RGB image, coarse depth $\hat{D}$ before refinement, and final depth output $D$ after refinement. $\ell_2$ loss computed between the groundtruth and inference result is presented at the top left corner. The coarse depth map is significantly refined by our deformable refinement module {\em with only a single pass}.}
  \vspace{-3mm}
\label{fig:comp_ref}
\end{figure*}

Our evaluation so far has validated the importance of deformable convolution for depth completion, in a way of refining the estimated coarse depth. Here we further investigate alternative strategies of exploiting deformable convolution, and show that our current approach is the most effective to employ it to improve depth completion performance, {\em especially when the input depth maps are of very high sparsity}. Figure~\ref{fig:ablation} illustrates several representative variants and Table~\ref{table:ablation} summarizes their performance. 

Specifically, (I) we start from a straightforward use of deformable convolution as \textbf{Variant 1}, where we directly apply the learned weights and offsets on the {\em input} sparse depth $X_d$ following DKN~\cite{kim2019deformable}. We use the same backbone as in DKN.  This method has been shown to achieve high performance on the NYU-v2 dataset~\cite{Silberman:ECCV12} where the density of the input depth maps is much higher than KITTI~\cite{kim2019deformable}. However, as shown in Table~\ref{table:ablation}, it achieves very poor performance with RMSE of $1130.47$ on KITTI. (II) We then upgrade \textbf{Variant 1 with a more powerful backbone} -- the one used in both PENet and our current model. Despite the observed improvement ($900.54$ v.s. $1130.47$ RMSE), it still fails to match competitive depth completion performance on KITTI. This shows that it is difficult to operate deformable convolution directly on {\em input depth of high sparsity}, irrespective of an improved feature extractor or backbone.

We thus conjecture that deformable kernel convolution requires a relatively high density for input depth maps to obtain good performance. Based on this assumption, we pre-process the depth maps to increase their density by using various interpolation techniques. We directly test the interpolated depth from the sparse depth maps against the groundtruth on the KITTI validation set, and report the results in Table~\ref{table:interpolation}. We observe that the nearest-neighbor interpolation, filling the empty depth values based on their nearest neighbors, achieves the best overall performance. 

Based on this result, (III) we further explore \textbf{Variant 2} (Figure~\ref{fig:ablation}), where we feed the nearest-neighbor interpolated depth into Variant 1 with a strong backbone. As expected, we observe an improvement ($875.75$ v.s. $900.54$ RMSE) in Table \ref{table:ablation}, which validates our assumption that the density of the input depth maps is crutial for the performance of deformable convolution. Compared with Variant 2, our model is substantially better under all the metrics, since we use a {\em learned} coarse depth map. Collectively, the comparisons with Variants 1 and 2 in Table~\ref{table:ablation} demonstrate the superiority of our strategy -- an intermediate coarse depth map is first generated under the guidance of an RGB image via a deep backbone; then, it gets improved by our deformable refinement module.

\subsection{Ablation Studies}
\label{subsec:ablation}

We conduct a variety of ablation studies to further demonstrate the performance improvement brought by our refinement module, and investigate the effect of different components and hyperparameter settings in our approach.

\textbf{Coarse-to-Fine Procedure:}
\label{subsubsec:coarse}
Figure~\ref{fig:comp_ref} visually compares the generated coarse depth $\hat{D}$ before our refinement module and the final depth map $D$. We also provide the $\ell_2$ loss computed between the groundtruth depth maps and the generated depth maps at the top left corner as in Section~\ref{subsubsec:visul}. We observe that the quality of the generated depth gets significantly improved by our refinement module. Depth values in many areas of the coarse depth maps get corrected, and the depth detail of objects also gets refined. We also provide the result of training and validating the backbone alone without our refinement module in Table \ref{table:hyperparameter} which shows obviously worse performance (835.37 v.s. 813.05 RMSE), highlighting the effectiveness of our refinement.

\textbf{Hyperparameter Settings:}
\begin{table}[t!]
\vspace{-2mm}
    \centering
    \caption{\small Ablation studies on hyperparameter settings on the KITTI validation set. Our current setting achieves the best performance.}
    \vspace{-3mm}
    \label{table:hyperparameter}
    \resizebox{0.47\textwidth}{!}
    {
        \begin{tabular}{c|cccc}
            \hline
            Model  & RMSE ($\downarrow$) & MAE ($\downarrow$) & iRMSE ($\downarrow$) & iMAE ($\downarrow$)\\\hline
            Backbone Only & 835.37 & 235.00 &2.47 &0.97\\\hdashline
             ReDC w/ staircase lr decay & 820.76 &228.31 &\textbf{2.30} &0.95\\
             ReDC w/ $\ell_2$ loss & 815.53 &227.35 &2.38 &0.95\\
             ReDC w/ $\ell_2$ loss + staircase  & 821.53 & 233.90 & 2.32 & 0.98\\
             ReDC $k=5$ & 850.34 &234.31 &\textbf{2.30} &0.95\\
            \rowcolor{lightgray} ReDC (Ours) &\textbf{813.05} &\textbf{218.30} &2.32 &\textbf{0.89}\\
            \hline
        \end{tabular}
    }
    \vspace{-6mm}
\end{table}
We study the impact of different hyperparameter settings in Table~\ref{table:hyperparameter}. (I) \textbf{Learning rate schedule:} Instead of decaying the learning rate in a cosine annealing fashion, we use a staircase schedule. Specifically, the learning rate gets decayed by $\{0.5, 0.2, 0.1\}$ at epoch $\{10, 15, 25\}$. The performance is not as good as cosine annealing ($820.76$ v.s. $813.05$ RMSE). (II) \textbf{Training objective:} The performance gets worse ($815.53$ v.s. $813.05$ RMSE), when the training objective is switched to only an $\ell_2$ loss instead of a combined $\ell_2$ and $\ell_1$ loss. The performance further drops, when we use both staircase decay and $\ell_2$ loss ($821.53$ v.s. $813.05$ RMSE). (III) \textbf{Deformable kernel size:} We investigate whether a larger deformable kernel $W$ would improve the completion performance. We observe that, when $k=5$, the performance gets significantly degraded ($850.34$ v.s. $813.05$ RMSE). These ablations show that our current hyperparameter setting achieves the best performance.

\section{CONCLUSION}
In this paper, we revisited deformable convolution and studied its best usage on depth completion with very sparse depth maps. We found that different ways of employing deformable convolution could lead to completely distinct performance, and it needs to be applied on a depth map with a relatively high density for optimal performance. We proposed a new, effective depth completion architecture that requires only a single pass through our deformable refinement module, which separates our model from previous ones which usually require a minimum of three passes to get good performance. We achieved competitive state-of-the-art level depth completion performance on the challenging KITTI dataset and surpassed previous work by a clear margin. We believe that this work will deepen the understanding of deformable convolution in the depth completion community.
\section*{ACKNOWLEDGMENT}
This work was supported in part by NSF Grant 2106825, NIFA Award 2020-67021-32799, the Jump ARCHES endowment, the NCSA Fellows program, the Inria/NYU collaboration, the Louis Vuitton/ENS chair on artificial intelligence and the French government under management of Agence Nationale de la Recherche as part of the Investissements d’avenir program, reference ANR19-P3IA0001 (PRAIRIE 3IA Institute). We thank Beomjun Kim and Bumsub Ham for providing their DKN code.

{\small
\bibliographystyle{IEEEtran}
\bibliography{IEEEfull}
}
\end{document}